\documentclass[10pt,twocolumn,letterpaper]{article}

\usepackage[pagenumbers]{cvpr} 

\usepackage[dvipsnames]{xcolor}

\usepackage[normalem]{ulem}
\usepackage{multirow}
\usepackage{colortbl}
\usepackage{hhline}
\usepackage{color}
\usepackage{tabularray}
\usepackage[accsupp]{axessibility}

\definecolor{cvprblue}{rgb}{0.21,0.49,0.74}
\usepackage[pagebackref,breaklinks,colorlinks,citecolor=cvprblue]{hyperref}

\def\paperID{15947} 
\def\confName{CVPR}
\def\confYear{2024}

\title{Text Prompt with Normality Guidance for Weakly Supervised Video Anomaly Detection}

\author{Zhiwei Yang$^1$, Jing Liu$^1$\thanks{Corresponding authors.}, Peng Wu$^2$\\ 
$^1$Guangzhou Institute of Technology, Xidian University, Guangzhou, China\\ 
$^2$School of Computer Science, Northwestern Polytechnical University, Xi'an, China\\
{\tt\small \{zwyang97, neouma\}@163.com, xdwupeng@gmail.com}
}

\begin{document}
\maketitle
\begin{abstract}
Weakly supervised video anomaly detection (WSVAD) is a challenging task. Generating fine-grained pseudo-labels based on weak-label and then self-training a classifier is currently a promising solution. However, since the existing methods use only RGB visual modality and the utilization of category text information is neglected, thus limiting the generation of more accurate pseudo-labels and affecting the performance of self-training. Inspired by the manual labeling process based on the event description, in this paper, we propose a novel pseudo-label generation and self-training framework based on \textbf{T}ext \textbf{P}rompt \textbf{w}ith \textbf{N}ormality \textbf{G}uidance (TPWNG) for WSVAD. Our idea is to transfer the rich language-visual knowledge of the contrastive language-image pre-training (CLIP) model for aligning the video event description text and corresponding video frames to generate pseudo-labels. Specifically, We first fine-tune the CLIP for domain adaptation by designing two ranking losses and a distributional inconsistency loss. Further, we propose a learnable text prompt mechanism with the assist of a normality visual prompt to further improve the matching accuracy of video event description text and video frames. Then, we design a pseudo-label generation module based on the normality guidance to infer reliable frame-level pseudo-labels. Finally, we introduce a temporal context self-adaptive learning module to learn the temporal dependencies of different video events more flexibly and accurately. Extensive experiments show that our method achieves state-of-the-art performance on two benchmark datasets, UCF-Crime and XD-Violence, demonstrating the effectiveness of our proposed method.
\end{abstract}

\section{Introduction}
\label{sec:intro}

Anomaly detection has been widely researched and applied in various fields, such as computer vision \cite{yang2023video-88, wu2023towards-102, zavrtanik2021draem-105, shao2022exploiting-110,yang2023slsg-111}, natural language processing \cite{bertero2017experience-104}, and intelligent optimization \cite{wang2021solving-103}. One of the most important research issues is the video anomaly detection (VAD). 
\begin{figure}[t]
  \centering
   \includegraphics[width=0.75\linewidth]{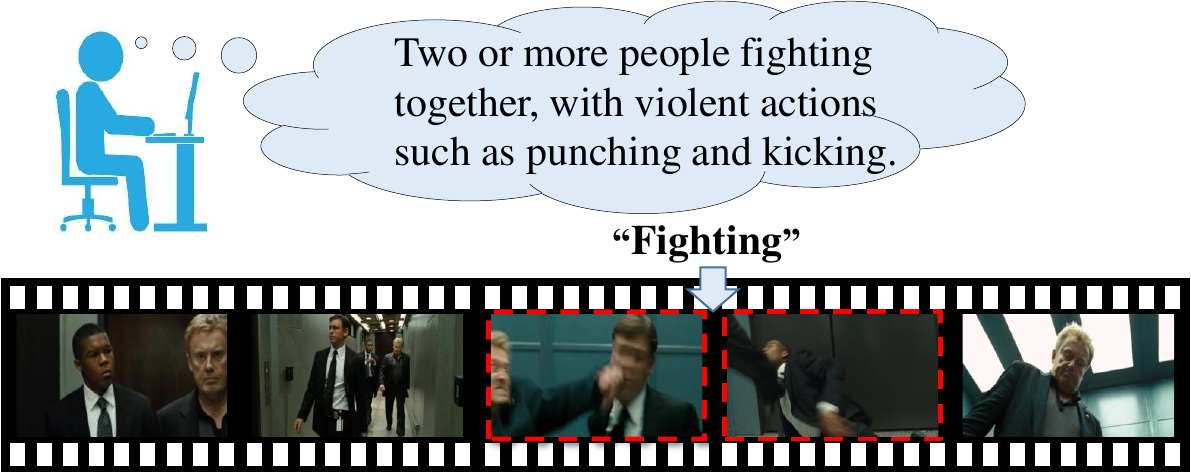}

   \caption{Illustration of the manual video frame labeling process.}
   \label{fig:shouyetu}
\end{figure}
The main purpose of VAD is to automatically identify events or behaviors in the video that are inconsistent with our expectations.

Due to the rarity of anomalous events and the difficulty of frame-level labeling, current VAD methods focus on semi-supervised \cite{liu2018future-9, park2020learning-45,lv2021learning-47} and weakly supervised \cite{ sultani2018real-67, zhang2019temporal-72, li2022self-64} paradigms. Semi-supervised VAD methods aim to learn normality patterns from normal data, and deviations from this pattern are considered as anomalies. However, due to the lack of discriminative anomaly information in the training phase, these models are often prone to overfitting, leading to poor performance in complex scenarios. Subsequently, weakly supervised video anomaly detection (WSVAD) methods came into prominence. WSVAD involves both normal and abnormal videos with video-level labels in the training phase, but the exact location of abnormal frames is unknown. Current WSVAD methods mainly include one-stage methods based on multi-instance learning (MIL) \cite{lv2021localizing-65,sultani2018real-67, tian2021weakly-68} and two-stage methods based on pseudo-label self-training \cite{zhong2019graph-73, feng2021mist-63, zhang2023exploiting-78, li2022self-64}. While the one-stage methods based on MIL show promising results, this paradigm tends to focus on video snippets with prominent anomalous features and suboptimal attention to minor anomalies, thus limiting its further performance improvement.

In contrast to the one-stage methods mentioned above, two-stage methods based on pseudo-label self-training generally use an off-the-shelf classifier or MIL to obtain initial pseudo-labels, and then train the classifier with further refined pseudo-labels. Because these methods train the classifier directly with the generated fine-grained pseudo-labels, they show great potential in performance. However, these methods still have two aspects that have not been considered: first, the generation of pseudo-labels is based only on visual modality and lacks the utilization of textual modality, which limits the accuracy and completeness of the generated pseudo-labels. Second, the mining of temporal dependencies among video frames is insufficient.

To further exploit the potential of pseudo-label-based self-training on WSVAD, we dedicate to investigating the two problems mentioned above in this paper.  \textbf{Our motivation for the first question} is that we explore how the textual modal information can be effectively utilized to assist in generating pseudo-labels. Recalling our manual process of video frame labeling, we mainly based on textual definitions of anomalous events, i.e., prior knowledge about anomalous events, to accurately locate the video frames. As illustrated in \cref{fig:shouyetu}, assuming that we need to annotate the abnormal video frames that contain “fighting” event, we will first associate the textual definition of “fighting” 
and then look for matching video frames, which is actually a process of text-image matching based on prior knowledge. Inspired by this process, we associate a highly popular and powerful contrastive language-image pre-training  (CLIP) \cite{radford2021learning-94} model to assist us in achieving this goal. On the one hand, the CLIP learns a large number of image-text pairs on the web, and thus has a highly rich prior knowledge; on the other hand, the CLIP is trained by comparative learning, which empowers it with excellent image-text alignment capabilities.
\textbf{For the second motivation}, because different video events have diverse durations, this leads to different ranges of temporal dependencies. Existing methods
either do not consider temporal dependencies or only consider dependencies within a fixed temporal range, leading to inadequate modeling of temporal dependencies. Therefore, in order to achieve more flexible and adequate modeling of temporal dependencies, we should investigate methods that can adaptively learn temporal dependencies of different lengths.

Based on the above two motivations, we propose a novel pseudo-label generation and self-training framework based on \textbf{T}ext \textbf{P}rompt \textbf{w}ith \textbf{N}ormality \textbf{G}uidance (TPWNG) for WSVAD. Our main idea is to utilize the CLIP model to match the textual descriptions of video events with the corresponding video frames, and then infer the pseudo-labels from match similarities. However, since the CLIP model is trained at the image-text level, it may suffer from domain bias and lacks the ability to learn temporal dependencies in videos. In order to better transfer the prior knowledge of CLIP to the WSVAD task, we first construct a contrastive learning framework by designing two ranking losses and a distributional inconsistency loss to fine-tune the CLIP model for domain adaptation under the weakly-supervised setting. To further improve the accuracy of aligning the descriptive text of video events with video frames, we employ learnable textual prompts to facilitate the text encoder of CLIP to generate more generalized textual embedding features. On this basis, we propose a normality visual prompt (NVP) mechanism to aid this process. In addition, because abnormal videos contain normal video frames as well, we design a pseudo-label generation (PLG) module based on normality guidance,
which can reduce the interference caused by individual normal video frames to the alignment of abnormal video frames, thus facilitating the obtaining of more accurate frame-level labels. 

Furthermore, to compensate for the lack of temporal relationship modeling in CLIP as well as to more flexible and adequately mine the temporal dependencies between video frames, we introduce a temporal context self-adaptive learning (TCSAL) module for temporal dependency modeling, inspired by the work \cite{sukhbaatar2019adaptive-95}. TCSAL allows the attention module in the Transformer to adaptively adjust the attention span according to the inputs by designing a temporal span adaptive learning mechanism. This can facilitate the model to capture the temporal dependencies of video events of different durations more accurately and flexibly.

Overall, our main contributions are summarized below:
\begin{itemize}
\item We propose a novel framework, i.e., TPWNG, to perform pseudo-label generation and self-training for WSVAD. TPWNG fine-tunes CLIP with the designed ranking loss and distributional inconsistency loss to transfer its strong text-image alignment capability to assist pseudo-label generation by means of the PLG module.
\item We design a learnable text prompt and normality visual prompt mechanism to further improve the alignment accuracy of video events description text and video frames.
\item We introduce a TCSAL module to learn the temporal dependencies of different video events more flexibly and accurately. To the best of our knowledge, we are the first to introduce the idea of self-adaptive learning of temporal context dependencies for VAD.
\item Extensive experiments have been conducted on two benchmark datasets, UCF-Crime and XD-Violence, where the excellent performance demonstrates the effectiveness of our method.
\end{itemize}

\section{Related Work}
\label{sec:Related Work}

\subsection{Video Anomaly Detection}
The VAD task has been widely focused and researched, and many methods have been proposed to solve this problem. According to different supervision modes, these methods can be mainly categorized into semi-supervised-based and weakly supervised-based VAD.

\textbf{Semi-supervised VAD.} Early researchers mainly used semi-supervised approaches to solve the VAD problem \cite{sabokrou2015real-4, zhai2016deep-5,gong2019memorizing-6,liu2018future-9,hasan2016learning-8,ye2019anopcn-27,luo2017revisit-40, zaheer2022stabilizing-87, yang2023video-88, slavic2022kalman-89, lee2019bman-90, yang2021bidirectional-109,cai2021appearance-49, yang2022dynamic-28, wu2019deep-55,wang2021robust-48}. In the semi-supervised setting, only normal data can be acquired in the training phase, which aims to build a model that can characterize normal behavioral patterns by learning normal data. During the testing phase, data that contradict with the normal patterns are considered anomalies. Common semi-supervised VAD methods mainly include one-class classifier-based \cite{xu2015learning-79, wu2019deep-55, sabokrou2018adversarially-80} and reconstruction \cite{hasan2016learning-8, xu2017detecting-24} or prediction errors-based methods \cite{liu2018future-9, yang2022dynamic-28}. For example, Xu et al. \cite{xu2017detecting-24} used multiple one-classifiers to predict anomaly scores based on appearance and motion features. Hasan et al. \cite{hasan2016learning-8} built a fully convolutional auto-encoder to learn regular patterns in the video. 
Liu et al. in \cite{liu2018future-9} proposed a novel video anomaly detection method that utilizes the U-Net architecture to predict future frames, where frames with large prediction errors are considered as anomalous.

\textbf{Weakly Supervised VAD.} Compared to semi-supervised VAD methods, WSVAD can utilize both normal and anomalous data with video-level labels in the training phase, but the exact frame location where the abnormal event occurred is unknown. In such a setting, the one-stage approaches based on MIL \cite{sultani2018real-67, tian2021weakly-68, wu2021learning-69, wu2020not-70, zaheer2020claws-71, cho2023look-75, zhou2023dual-76, chen2023mgfn-77, liu2023distilling-92,cho2023look-93, lv2021localizing-65, sapkota2022bayesian-66} and the two-stage approaches based on pseudo-labels self-training  \cite{zhong2019graph-73, feng2021mist-63, zhang2023exploiting-78, li2022self-64} are the two prevailing approaches. For example, Sultani et al. \cite{sultani2018real-67} first proposed a deep MIL ranking framework for VAD, where they considered anomalous and normal videos as positive and negative bags, respectively, and the snippets in the videos are considered as instances. Then a ranking loss is used to constrain the snippets with the highest anomaly scores in the positive and negative bags to stay away from each other. Later, many variants of the method were proposed on this basis. For example, Tian et al. \cite{tian2021weakly-68} proposed a top-k MIL based VAD method with robust temporal feature magnitude learning. 



However, these one-stage methods generally use a MIL framework, which leads to models that tend to focus only on the most significant anomalous snippets while ignoring nontrivial anomalous snippets. A two-stage approach based on pseudo-label self-training provides a relatively more promising solution. The two-stage approach first generates initial pseudo-labels using MIL or an off-the-shelf classifier and then refines the labels before using them for supervised training of the classifier. For example, Zhong et al. in \cite{zhong2019graph-73} reformulated the WSVAD problem as a supervised learning task under noisy labels obtained by an off-the-shelf video classifier. Feng et al. in \cite{feng2021mist-63} introduced a multiple instance pseudo label generator that produces more reliable pseudo labels for fine-tuning a task-specific feature encoder with self-training mechanism. Zhang et al. in \cite{zhang2023exploiting-78} exploited completeness and uncertainty properties to enhance pseudo labels for effective self-training. However, all these existing methods only generate pseudo-labels based on visual unimodal information and lack the utilization of textual modal. Therefore, in this paper, we endeavor to combine both visual and textual modal information in order to generate more accurate and complete pseudo-labels for self-training of the classifier.
\subsection{Large Vision-Language Models}
Recently, there has been an emergence of large vision-language models that learn the interconnections between visual and textual modalities by pre-training on large-scale datasets. Among these methods, the CLIP demonstrates unprecedented performance in many visual-language downstream tasks, e.g. image classification \cite{zhou2022learning-97}, object detection \cite{zhou2022detecting-98}, semantic segmentation \cite{lin2023clip-99} and so on. The CLIP model has recently been successfully extended to the video domain as well. VideoCLIP \cite{xu2021videoclip-100} is proposed to align video and textual representations by contrasting temporally overlapping video-text pairs with mined hard negatives. ActionCLIP \cite{wang2021actionclip-101} formulated the action recognition task as a multimodal learning problem rather than a traditional unimodal classification task. However, there are fewer attempts to utilize CLIP models to solve VAD tasks. Joo et al. in \cite{joo2023clip-96} simply utilizes CLIP's image encoder for extracting more discriminative visual features and does not use textual information. Wu et al. \cite{wu2023vadclip-107}, Zanella et al. \cite{zanella2023delving-108} mainly use textual features from CLIP to enhance the expressiveness of the overall features, followed by MIL-based anomaly classifier learning. The major difference with the above works is that our method is the first to utilize the textual features encoded by the CLIP text encoder in conjunction with the visual features to generate pseudo-labels, and then employ a supervised approach to train an anomaly classifier.

\section{Method}
In this section, we first present the definition of the WSVAD task, then introduce the overall architecture of our proposed method, and subsequently elaborate on the details of each module and the execution process.
\begin{figure*}[t]
  \centering
   \includegraphics[width=0.80\linewidth]{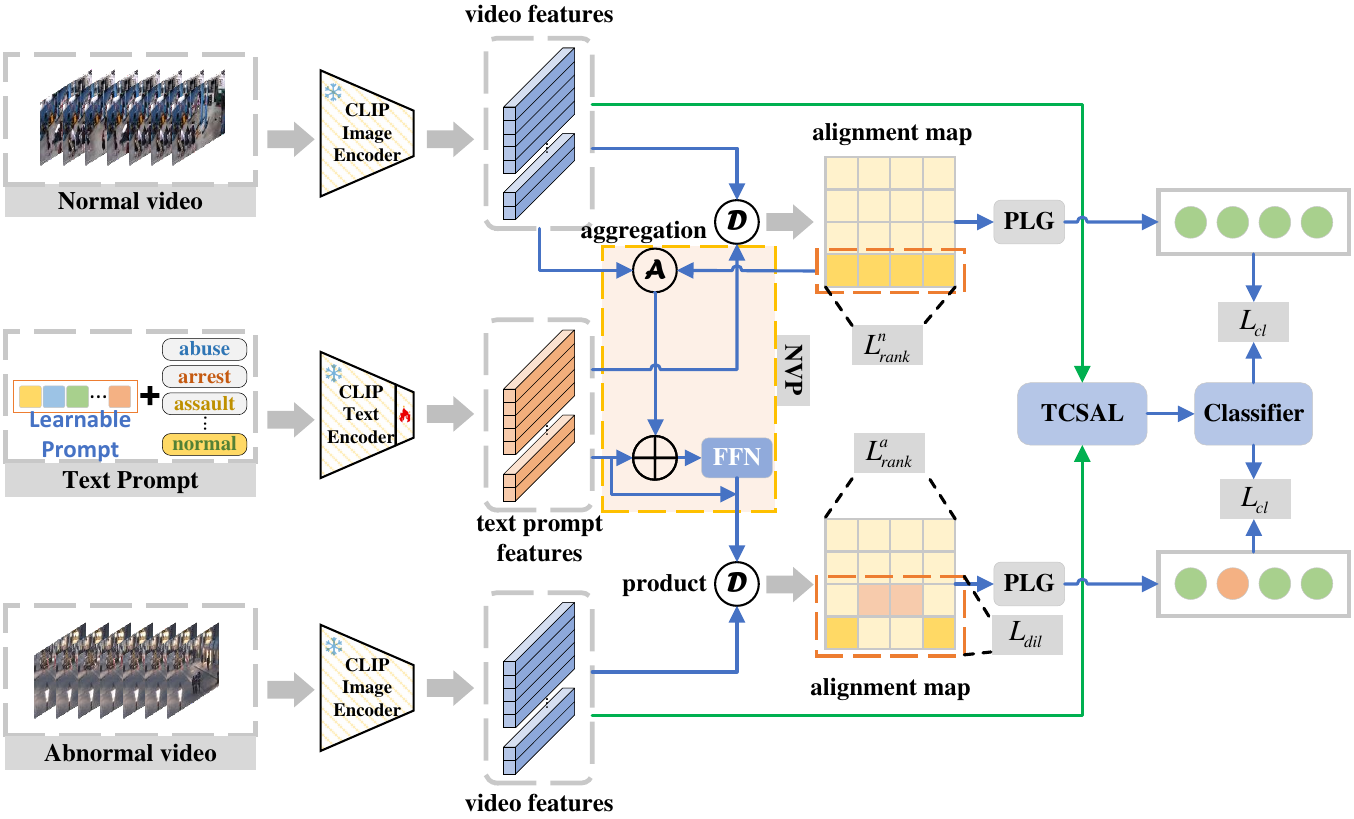}

   \caption{The overall architecture of our proposed TPWNG.}
   \label{fig:main}
\end{figure*}
\subsection{Overall Architecture}
Formally, we first define sets ${D^a}=\{(v_i^a,y_i)\}_{i = 1}^M$ and ${D^n}=\{ (v_i^n,y_i)\} _{i=1}^M$ containing $M$ abnormal and normal videos with ground-truth labels, respectively. For each $v_i^a$, it is labeled $y_i = 1$, indicating that this video contains at least one anomalous video frame, but the exact location of the anomalous frame is unknown. For each $v_i^n$, it is labeled $y_i=0$, indicating that this video consists entirely of normal frames. With this setting, WSVAD task is to utilize coarse-grained video-level labels to enable a classifier to learn to predict fine-grained frame-level anomaly scores. 

\cref{fig:main} illustrates the overall pipeline of our approach. Normal and abnormal video along with learnable category prompt text are encoded as feature embedding by the image encoder and text encoder of CLIP, respectively. Then, the text encoder of CLIP is encouraged by fine-tuning it to produce textual feature embedding of video event categories that accurately match anomalous or normal video frames, and the NVP assists in this process. Meanwhile, the image features feed the TCSAL module to perform self-adaptive learning of temporal dependencies. Finally, a video frame classifier is trained to predict anomaly scores under the supervision of pseudo-labels obtained by the PLG module.

\subsection{Text and Normality Visual Prompt}
\textbf{Learnable Text Prompt.} Constructing textual prompts that can accurately describe various video event categories is a prerequisite for realizing the alignment of text and corresponding video frames. However, it is impractical to manually define description texts that can completely characterize anomalous events in all different scenarios. Therefore, inspired by CoOp \cite{zhou2022learning-97}, we employ a learnable text prompt mechanism to adaptively learn representative video event text prompts to align the corresponding video frames. Specifically, we construct a learnable prompt template, which adds $l$ learnable prompt vectors in front of the tokenized category name, as follows:
\begin{equation}
 {p_{label}} = ({\partial _1},\;...,\;{\partial _l},\;Tokenizer(label)),
  \label{eq:eq1}
\end{equation}
where ${\partial _l}$ denotes the $l{\rm{-th}}$ prompt vector. Tokenizer is converting original category labels, i.e., ``fighting", ``accident", \ldots, ``normal", etc., into class tokens by means of CLIP tokenizer. Then, we add the corresponding location information $pos$ to the learnable prompts and then feed it to the CLIP text encoder ${\zeta _{text}}$ to get the feature embedding ${T_{label}} \in {\mathbb{R}^D}$ of the video event description text as follows:
\begin{equation}
{T_{label}} = {\zeta _{text}}({p_{label}} \oplus \;pos),
  \label{eq:eq2}
\end{equation}
Finally, we compute all video event categories according to \cref{eq:eq1,eq:eq2} to obtain the video event description text embedding set $E = \{ T_1^a,\;T_2^a,\;...,\;T_{k - 1}^a,\;T_k^n\} $, where $\{T_i^a\}_{i=1}^{k-1}$ denotes the description text embedding of preceding $k - 1$ abnormal events and $T_k^n$ denotes the description text embedding of normal events.

\textbf{Normality Visual Prompt.} 
For an anomalous video, which contains both anomalous and normal frames, our core task is to infer pseudo-labels from the match similarities between the description text of the anomalous events and the video frames. However, this process is susceptible to interference from normal frames in the anomalous video because they have a similar background to the anomalous frames. To minimize this interference, we propose a NVP mechanism. NVP is used to assist the normal event description text to more accurately align normal frames in the abnormal video, and thus indirectly assist the description text of abnormal event to align abnormal video frames in the abnormal video by means of the distribution inconsistency loss that will be introduced in \cref{sec:obj}. Specifically, we first compute the match similarities $S_{i,\;k}^{nn} \in {\mathbb{R}^F}$ between the description text embedding of normal event and the video frame features in the normal video. Then, the match similarities after softmax operation are used as weights to aggregate normal video frame features to obtain NVP ${Q_i} \in {\mathbb{R}^D}$. The formulas are represented as follows:
\begin{equation}
S_{i,\;k}^{nn} = X_i^n{(T_k^n)^ \top }, {Q_i} = softmax({(S_{i,\;k}^{nn})^ \top })X_i^n,
  \label{eq:eq3}
\end{equation}
where $X_i^n \in {\mathbb{R}^{F \times D}}$ denotes the visual features of the normal video $v_i^n$ obtained by the CLIP image encoder, where $F$ and $D$ denote the number of video frames and feature dimensions, respectively. Then, we concatenate ${Q_i}$ and $T_k^n$ in the feature dimension and feed an FFN layer with skip connections to obtain the enhanced description text embedding $\dot T_k^n$ of normal events. The formula is represented as follows:
\begin{equation}
\dot T_k^n = FFN((T_k^n \cup {Q_i})) + T_k^n.
  \label{eq:eq5}
\end{equation}
\subsection{Pseudo Label Generation Module}
In this subsection, we detail how to generate frame-level pseudo labels. For a normal video, we can directly get the frame-level pseudo-labels, i.e., for a $v_i^n = \{ {I_j}\} _{j = 1}^F$ containing $F$ normal frames, it corresponds to a label set $\{\gamma_{i,\;j}^n=0\}_{j = 1}^F$. Our main goal is to infer the pseudo-labels for anomalous videos that contain both anomalous and normal frames. To this end, we propose a PLG module for inferring accurate pseudo-labels based on the normality guidance. PLG module infers frame-level pseudo-labels by incorporating the match similarities between the description text of the normal event and the abnormal video as a guide into the match similarities between the description text of the corresponding abnormal event and the abnormal video. 

Specifically, we first compute the match similarities $S_{i,\;k}^{an} = X_i^a{(\dot T_k^n)^ \top }$ between normal event description text embedding enhanced with NVP and anomalous video features, where $X_i^a\in{\mathbb{R}^{F \times D}}$ denotes the visual features of the anomalous video $v_i^a$ obtained by the CLIP image encoder. Similarly, we compute the match similarities $S_{i,\;\tau }^{aa} = X_i^a{(T_\tau ^a)^ \top }$ between the description text embedding $T_\tau ^a$ of the corresponding $\tau{\text{-th}}\;(1 \leqslant \tau  \leqslant k - 1)$ real anomaly category and the anomaly video features $X_i^a$.

Theoretically, for $S_{i,\;\tau }^{aa}$, it should have high match similarities corresponding to abnormal frames and low match similarities for normal frames. But it may be interfered by normal frames from the same video having the same background. To reduce the interference of normal frames, we infer pseudo-labels by incorporating the matching similarity corresponding to the description text of normal events with certain weights as a guide into the matching similarity of the description text of corresponding real abnormal events. Specifically, we first perform a normalization and fusion operation on $S_{i,\;\tau }^{aa}$ and $S_{i,\;k}^{an}$ as follows:
\begin{equation}
{\psi _i} = \alpha \tilde S_{i,\;k}^{an} + (1 - \alpha )(1 - \tilde S_{i,\;\tau }^{aa}),
  \label{eq:eq8}
\end{equation}
where $\tilde  * $ denotes the normalization operation and $\alpha $ denotes the guidance weight. After obtaining ${\psi _i}$, we similarly perform a normalization operation on it to obtain ${\tilde \psi _i}$. Then, we set a threshold $\theta $ on ${\tilde \psi _i}$ to obtain the frame-level pseudo-labels in the anomalous video as follows:
\begin{equation}
\gamma _{i,\;j}^a = \left\{ \begin{array}{l}
1,\;{\tilde\psi _{i,j}} \ge \theta ;\\
0,\;{\tilde\psi _{i,j}} < \theta ,
\end{array} \right.i = 1,2,...,M;\;j = 1,2,...,F
  \label{eq:eq9}
\end{equation}
where $\gamma _{i,\;j}^a$ denotes the pseudo-label of the $j{\text{-th}}$ frame in the $i{\text{-th}}$ anomaly video. Finally, we combine the frame-level pseudo-labels $\gamma _{i,\;j}^n$ and $\gamma _{i,\;j}^a$ of normal and anomalous videos to get the total pseudo-label set $\{ {\gamma _{i,\;j}}\} _{j = 1}^F$.

\subsection{Temporal Context Self-adaptive Learning}
To adaptively adjust the learning range of temporal relationship based on the input video data, inspired by the work \cite{sukhbaatar2019adaptive-95}, we introduce a TCSAL module. The backbone of TCSAL is the transformer-encoder, but unlike the original transformer, the spanning range of attention is controlled by a soft mask function  ${\chi _z}$ for each self-attention head at each layer.  ${\chi _z}$ is a piecewise function mapping a distance to a value between [0, 1] as follows:
\begin{equation}
{\chi _z}(h) = \min \left[ {\max \left[ {\frac{1}{R}(R + z - h),\;0} \right],\;1} \right],
  \label{eq:eq10}
\end{equation}
where $h$ represents the distance between the current $t{\text{-th}}$ frame in a video and the $r{\rm{-th}}$ ($r \in [1,t-1]$) frame in the past temporal range. $R$ is a hyperparameter used to control the softness. $z$ is a learnable parameter that is adaptively tuned with the input as follows: 
\begin{equation}
z = F\sigma ({C^ \top }X + b),
  \label{eq:eq11}
\end{equation}
here $\sigma$ represents the sigmoid operation, $C$ and $b$ are learnable parameters during model training. 
With the soft mask function ${\chi _z}$, the corresponding attention weights ${\omega _{t,\;r}}$ is computed within this mask, i.e.,
\begin{equation}
{\omega _{t,\;r}} = \frac{{{\chi _z}(t - r)\exp ({\beta _{t,r}})}}{{\sum\nolimits_{q = 1}^{t - 1} {{\chi _z}(t - q)\exp ({\beta _{t,q}})} }},
  \label{eq:eq12}
\end{equation}
here ${\beta _{t,r}}$ denotes the dot product output of the \textit{Query} corresponding to the $t{\text{-th}}$ frame in a video with the  \textit{Key} corresponding to the $r{\rm{-th}}$ frame in the past. Under the control of ${\chi _z}$, the self-attention heads will be able to adaptively adjust the self-attention span range according to the input.

Finally, the video features after temporal context adaptive learning are fed into a classifier to predict the frame-level abnormality scores $\{ {\eta _{i,\;j}}\} _{j = 1}^F$.
\subsection{Objective Function}
\label{sec:obj}
First, we fine-tune the CLIP text encoder. For a normal video, we further compute the match similarities set $\varphi _i^{na} = \{ S_{i,\;\tau }^{na} = X_i^n{(T_\tau ^a)^ \top }\left| {1 \leqslant \tau  \leqslant k - 1} \right.\} $ between the description texts of the other $k - 1$ anomalous events and the normal frames. We expect that the maximum in the similarity set $\varphi _i^{na}$ should be as small as possible while the maximum in $S_{i,\;k}^{nn}$ should be as large as possible. Thus, we design the following ranking loss for constraints:
\begin{equation}
L_{rank}^n = \max (0,\;1 - \max (S_{i,\;k}^{nn}) + \max (\max (\varphi _i^{na})).
  \label{eq:eq13}
\end{equation}
For an anomalous video, we first calculate the similarities $S_{i,\;k}^{an} = X_i^a{(\dot T_k^n)^ \top }$ between the description text embedding of normal event and the anomalous video features, the similarity $S_{i,\;\tau }^{aa} = X_i^a{(T_\tau ^a)^ \top }$ between the description text embedding of the $\tau {\text{-th}}\;(1 \leqslant \tau  \leqslant k - 1)$ real anomalous event category and the anomalous video features, and the similarity set  $\varphi _i^{aa} = \{ S_{i,\;g }^{aa} = X_i^a{(T_g^a)^ \top }\left| {1 \leqslant g \leqslant k - 1,\;g \ne \tau \;} \right.\} $ between the description text embedding of other $k - 2$ anomalous event categories and the anomalous video features, respectively. We expect that the maximum value in $S_{i,\;k}^{an}$ should be greater than the maximum value in $\varphi _i^{aa}$. Similarly, the maximum value in $S_{i,\;\tau }^{aa}$ should be greater than the maximum value in $\varphi _i^{aa}$. In short, it means that we expect that the description texts of real abnormal and normal events should match the abnormal and normal frames in the abnormal video with the highest possible similarity, respectively. Thus, the ranking loss for anomalous videos is designed as follows:
\begin{equation}
\begin{array}{l}
\begin{array}{l}
L_{rank}^a = \max (0,\;1 - \max (S_{i,\;k}^{an}) + \max (\max (\varphi _i^{aa}))) + \\
\;\;\;\;\;\;\max (0,\;1 - \max (S_{i,\;\tau }^{aa}) + \max (\max (\varphi _i^{aa}))).
\end{array}
\end{array}
  \label{eq:eq14}
\end{equation}
In addition, to further ensure that the description texts of real abnormal events and normal events can accurately align the abnormal and normal video frames in the abnormal video, respectively, we design a distribution inconsistency loss (DIL). DIL is used to constrain the similarities between the description text of the real abnormal event and the video frames to be inconsistent with the similarity distribution between the description text of the normal event and the video frames. We use cosine similarity to perform this loss:
\begin{equation}
{L_{dil}} = \frac{1}{{MF}}\sum\limits_{i = 1}^M {\sum\limits_{j = 1}^F {\frac{{\tilde S_{i,\;j,\;\tau }^{aa} \cdot \tilde S_{i,\;j,\;k}^{an}}}{{{{\left\| {\tilde S_{i,\;j,\;\tau }^{aa}} \right\|}_2} \cdot {{\left\| {\tilde S_{i,\;j,\;k}^{an}} \right\|}_2}}}} }.
  \label{eq:eq15}
\end{equation}
Then, following the work \cite{sultani2018real-67}, in order to make the generated pseudo-labels satisfy sparsity and smoothing in temporal order, we impose sparsity and smoothing constraints, ${L_{sp}}{\text{ = }}\sum\nolimits_{j = 1}^F {{{(\tilde S_{i,\;j,\;\tau }^{aa} - \tilde S_{i,\;j + 1,\;\tau }^{aa})}^2}} $, ${L_{sm}} = \sum\nolimits_{j = 1}^F {\tilde S_{i,\;j,\;\tau }^{aa}} $, on the similarity vectors $\tilde S_{i,\;\tau }^{aa}$. 

Then, we calculate the binary cross-entropy between the anomaly score ${\eta}_{i, j}$ predicted by the classifier and the pseudo-label ${\gamma _{i,\;j}}$ as the classification loss:
\begin{equation}
{L_{cl}} = -\frac{1}{{MF}}\sum\limits_{i = 1}^M {\sum\limits_{j = 1}^F {[{\eta}_{i, j} \log ({\gamma}_{i, j} ) \\
+ (1 - {\eta}_{i, j} )\log (1 - {\gamma}_{i, j} )]} }.
  \label{eq:eq16}
\end{equation}


The final overall objective function balanced by $\lambda _1$ and ${\lambda _2}$ is designed as follows:
\begin{equation}
{L_{all}} = L_{rank}^n + L_{rank}^a + {L_{dil}} + {L_{cl}} + {\lambda _1}{L_{sp}} + {\lambda _2}{L_{sm}}.
  \label{eq:eq17}
\end{equation}





\section{Experiments}
\subsection{Datasets and Evaluation Metrics}
\textbf{Datasets.} We conduct extensive experiments on two benchmark datasets, UCF-Crime \cite{sultani2018real-67} and XD-Violence \cite{wu2020not-70}. \textbf{UCF-Crime} is a large-scale real scene dataset for WSVAD. UCF-Crime duration is 128 hours in total and contains 1900 surveillance videos covering 13 anomaly event categories, of which 1610 videos with video-level labels are used for training and 290 videos with frame-level labels are used for testing.  \textbf{XD-Violence} is a large-scale violence detection dataset collected from movies, online videos, surveillance videos, CCTVS, etc. XD-Violence lasts 217 hours and contains 4754 videos covering 6 anomaly event categories, of which 3954 training videos with video-level labels and 800 test videos with frame-level labels.

\textbf{Evaluation Metrics.} Following the previous methods \cite{feng2021mist-63,sultani2018real-67}, for the UCF-Crime dataset, we measure the performance of our method using the area under the curve (AUC) of the frame-level receiver operating characteristics (ROC). Similarly, for the XD-Violence dataset, we follow the evaluation criterion of average precision (AP) suggested by the work \cite{wu2020not-70} to measure the effectiveness of our method.

\subsection{Implementation Details}

The image and text encoders in our method use a pre-trained CLIP (VIT-B/16), in which both the image and text encoders are kept frozen, except for the text encoder where the final projection layer is unfrozen for fine-tuning. The feature dimension $D$ is 512. FFN is a standard block from Transformer. The length $l$ of the learnable sequence in the text prompt is set to 8. The normality guidance weight $\alpha $ is set to 0.2 for both the UCF-Crime and XD-Violence datasets. The pseudo-labels generation threshold $\theta $ is set to 0.55 and 0.35 for the UCF-Crime and XD-Violence datasets, respectively. The parameter $R$ used to control the softness of the soft mask function is set to 256. The sparse loss and smoothing loss weights are set to ${\lambda _1} = 0.1$ and ${\lambda _2} = 0.01$. Please refer to the supplementary materials for more details on implementation.


\begin{table}
\centering
\arrayrulecolor{black}
\begin{tabular}{c|l|c|c} 
\arrayrulecolor{black}\hline
                                              & Methods                     & UCF (AUC)      & XD (AP)         \\ 
\cline{1-3}\arrayrulecolor{black}\cline{4-4}
\multirow{13}{*}{Weakly}                                 & Sultani et al.\cite{sultani2018real-67} \textbf{} & 77.92\%          & 73.20\%           \\
                                                         & GCN \cite{zhong2019graph-73}                    & 82.12\%           & -               \\
                                                         & HL-Net \cite{wu2020not-70}                 & 82.44\%           & 73.67\%           \\
                                                         & CLAWS \cite{zaheer2020claws-71}                  & 82.30\%          & -     \\
                                                         & MIST \cite{feng2021mist-63}                   & 82.30\%          & -               \\
                                                         & RTFM \cite{tian2021weakly-68}                    & 84.30\%          & 77.81\%           \\
                                                         & CRFD \cite{wu2021learning-69}                  & 84.89\%          & 75.90\%           \\
                                                         & GCL \cite{zaheer2022generative-84}                    & 79.84\%\          & -               \\
                                                         & MSL \cite{li2022self-64}                    & 85.62\%          & 78.58\%           \\
                                                         & MGFN \cite{chen2023mgfn-77}                   & 86.67\%          & 80.11\%           \\
                                                         & Zhang et al.\cite{zhang2023exploiting-78}            & 86.22\%          & 78.74\%           \\
                                                         & UR-DMU \cite{zhou2023dual-76}                     & 86.97\%          & 81.66\%           \\
                                                         & CLIP-TSA \cite{joo2023clip-96}                  & \uline{87.58\%}  & \uline{82.17\%}   \\ 
\hhline{>{\arrayrulecolor{black}}--->{\arrayrulecolor{black}}-}
\rowcolor[rgb]{0.906,0.902,0.902} \multicolumn{1}{l|}{~} & \textbf{Ours}               & \textbf{87.79\%} & \textbf{83.68\%}  \\
\hhline{>{\arrayrulecolor{black}}--->{\arrayrulecolor{black}}-}
\end{tabular}
\arrayrulecolor{black}
\caption{AUC and AP on UCF-Crime and XD-Violence dataset.}
\label{tab:tap1}
\end{table}

\subsection{Comparison with State-of-the-art Methods}
We compare the performance on the UCF-Crime and XD-Violence datasets with the current state-of-the-art (SOTA) methods in \cref{tab:tap1}. As can be observed from the table, our method achieves a new SOTA on both the UCF-Crime and XD-Violence datasets. Specifically, for the UCF-Crime dataset, our method outperforms the current SOTA method CLIP-TSA \cite{joo2023clip-96} by 0.21\%, which is not a trivial improvement for the challenging WSVAD task. 
\begin{figure*}[t]
  \centering
   \includegraphics[width=0.95\linewidth]{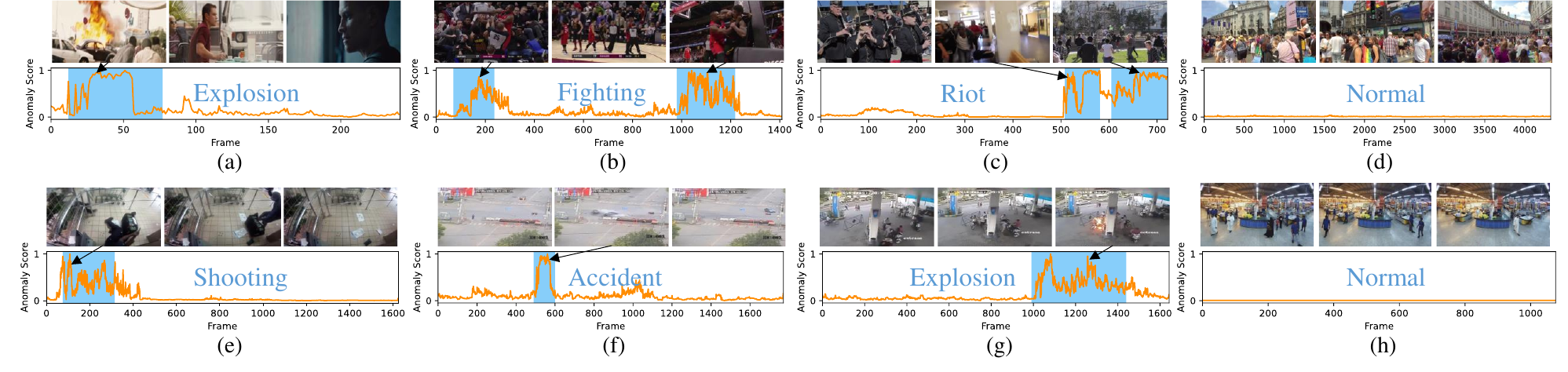}
   \caption{Anomaly score curves of several test samples on the UCF-Crime and XD-Violence dataset.}
   \label{fig:AS}
\end{figure*}
Most importantly, compared to methods MIST \cite{feng2021mist-63} and Zhang et al. \cite{zhang2023exploiting-78} similar to ours that also use pseudo-label-based self-training, our method significantly outperforms them by 5.49\% and 1.57\%, respectively. This fully demonstrates that our proposed pseudo-label generation and self-training framework is vastly superior to the above two approaches. This also indicates that transferring visual language multimodal associations through CLIP is conducive to generating more accurate pseudo-labels compared to merely utilizing unimodal visual information. For the XD-Violence dataset, our method also surpasses the current optimal method CLIP-TSA \cite{joo2023clip-96} by 1.52\%. Compared to a similar pseudo-label-based self-training method Zhang et al. \cite{zhang2023exploiting-78}, our method also outperforms it by 4.94\%. The consistent superior performance on two large-scale real datasets strongly demonstrates the effectiveness of our method. This also shows the extraordinary potential of the pseudo-label based self-training scheme, if accurate pseudo-labels can be generated utilizing multiple modality information.

\subsection{Ablation Studies}
We conduct ablation experiments in this subsection to analyze the effectiveness of each component of our framework.

\textbf{Effectiveness of Normal Visual Prompt.} To verify the validity of NVP, we execute three comparison experiments: without NVP, with NVP based on frame averaging (NVP-FA), and with NVP based on match similarities aggregation (NVP-AS).
\begin{table}
\centering
\arrayrulecolor{black}
\begin{tabular}{c!{\color{black}\vrule}c!{\color{black}\vrule}c} 
\hline
~        & UCF-Crime (AUC) & XD-Violence (AP)  \\ 
\cline{1-2}\arrayrulecolor{black}\cline{3-3}
w/o NVP  & 85.25\%           & 81.58\%             \\ 
\cline{1-1}\arrayrulecolor{black}\cline{2-2}\arrayrulecolor{black}\cline{3-3}
w NVP-FA & 87.32\%           & 83.13\%             \\ 
\cline{1-1}\arrayrulecolor{black}\cline{2-2}\arrayrulecolor{black}\cline{3-3}
w NVP-AS & 87.79\%           & 83.68\%             \\
\arrayrulecolor{black}\cline{1-2}\arrayrulecolor{black}\cline{3-3}
\end{tabular}
\arrayrulecolor{black}
\caption{The AUC and AP  of our method on the UCF and XD datasets without NVP, with NVP-FA, and with NVP-AS.}
\label{tab:tap2}
\end{table}
As can be seen from the results in \cref{tab:tap2}, in the absence of NVP, the performance of our method on the UCF-Crime and XD-Violence datasets decreases by 2.54\% and 2.10\% compared to with an NVP-AS, respectively. NVP-AS boosts the performance of the method by 0.47\% and 0.55\% more compared to NVP-FA on UCF-Crime and XD-Violence datasets, respectively. This reveals two facts: first, NVP can help the text embedding to better match normal frames in anomalous videos, which indirectly aids in generating more accurate pseudo-labels in cooperation with the DIL and the normality guidance mechanism. Second, the NVP-AS can effectively reduce the interference of some noise snippets (e.g., prologue, perspective switching, etc.) in normal videos compared to the NVP-FA approach, thus obtaining a purer NVP.

 %
 
\textbf{Effectiveness of the Normality Guidance.} In the pseudo-label generation module, instead of inferring pseudo-labels directly based on the similarity between the corresponding abnormal event description text and the abnormal video, we incorporate guidance from the match similarities of the normal event description text counterparts, aiming to reduce the interference of partially noisy video frames and generate more accurate pseudo-labels. To verify the contribution of the normality guidance, we compare the impact of the pseudo-label generation module on the performance of our method with and without normal guidance (NG), respectively.
As can be observed from \cref{tab:tap3}, when our method is equipped with normal guidance, the performance rises by 1.96\% and 2.36\% on the UCF-Crime and XD-Violence datasets, respectively. This validates the effectiveness of the normality guidance.

\begin{table}
\centering
\arrayrulecolor{black}
\begin{tabular}{c!{\color{black}\vrule}c!{\color{black}\vrule}c} 
\hline
~       & UCF-Crime (AUC) & XD-Violence (AP)  \\ 
\cline{1-2}\arrayrulecolor{black}\cline{3-3}
w/o NG & 85.83\%         & 81.32\%           \\ 
\arrayrulecolor{black}\cline{1-2}\arrayrulecolor{black}\cline{3-3}
w NG   & 87.79\%         & 83.68\%           \\
\arrayrulecolor{black}\cline{1-2}\arrayrulecolor{black}\cline{3-3}
\end{tabular}
\arrayrulecolor{black}
\caption{The AUC and AP of our method on the UCF and XD datasets with NG and without NG.}
\label{tab:tap3}
\end{table}

\textbf{Effectiveness of TCSAL.}
To analyze the effectiveness of TCSAL module, we conduct comparative experiments with the Transformer-encoder (TF-encoder) module in \cite{vaswani2017attention-106}, MTN module in \cite{tian2021weakly-68}, and GL-MHSA module in \cite{zhou2023dual-76} by replacing the temporal learning module in our framework with each of these three modules. From \cref{tab:tap4}, it can be observed that the TF-encoder module has the lowest performance, which is understandable since the global self-attention computation way makes it neglect to pay attention to the local temporal information. Both MTN and GL-MHSA outperform TF-encoder with comparable performance. Our introduced TCSAL module achieved the best performance on both datasets. This indicates that adopting the mechanism of self-attention span range adaptive learning enables the temporal learning module to self-adapt to the inputs of videos with different event lengths, achieving more accurate modeling of temporal dependencies while weakening the interference of other non-relevant temporal information in the non-event span range.
\subsection{Qualitative Results}
We show the anomalous scores of our method on several test videos in \cref{fig:AS}. It can be obviously noticed that there is a steep rise in the anomaly scores when various anomalous events occur, and as the anomalous events end, the anomaly scores fall back to the lower range rapidly. For normal events, our method gives a lower abnormal score. This intuitively demonstrates that our method has good sensitivity to abnormal events and can accurately and timely detect the occurrence of abnormal events while maintaining a low abnormal score prediction for normal events.

\begin{table}
\centering
\arrayrulecolor{black}
\begin{tabular}{l!{\color{black}\vrule}c!{\color{black}\vrule}c} 
\hline
\multicolumn{1}{c!{\color{black}\vrule}}{~} & UCF (AUC) & XD (AP)  \\ 
\cline{1-2}\arrayrulecolor{black}\cline{3-3}
w TF-encoder                                & 85.12\%         & 80.02\%           \\ 
\arrayrulecolor{black}\cline{1-2}\arrayrulecolor{black}\cline{3-3}
w MTN                                       & 86.22\%         & 81.02\%           \\ 
\arrayrulecolor{black}\cline{1-2}\arrayrulecolor{black}\cline{3-3}
w GL-MHSA                                   & 86.43\%         & 81.23\%           \\ 
\arrayrulecolor{black}\cline{1-2}\arrayrulecolor{black}\cline{3-3}
w TCSAL                                     & 87.79\%         & 83.68\%           \\
\arrayrulecolor{black}\cline{1-2}\arrayrulecolor{black}\cline{3-3}
\end{tabular}
\arrayrulecolor{black}
\caption{The AUC and AP of our method on the UCF and XD datasets with different temporal modules.}
\label{tab:tap4}
\end{table}


\begin{table}
\centering
\begin{tabular}{cccc|c|c} 
\hline
\multicolumn{4}{c|}{Loss term} & \multicolumn{2}{c}{Dataset}  \\ 
\hline
 $bs$         & $L_{rank}^n$ & $L_{rank}^a$ & ${L_{dil}}$ & UCF (AUC) & XD (AP) \\
\hline
$\checkmark$         &   &   &   & 77.12\%   & 73.32\% \\
$\checkmark$         & $\checkmark$ &   &   & 81.34\%   & 78.67\% \\
$\checkmark$         &   & $\checkmark$ &   & 84.45\%   & 81.56\% \\
$\checkmark$         &   &   & $\checkmark$ & 82.47\%   & 79.96\% \\
$\checkmark$         & $\checkmark$ & $\checkmark$ & $\checkmark$ & 87.79\%   & 83.68\% \\
\hline
\end{tabular}
\caption{ Comparison of the AUC and AP of our method with different loss terms on the UCF-Crime and XD-Violence datasets. "bs" indicates that ${L_{cl}}$, ${L_{sp}}$, ${L_{sm}}$ three loss functions are used.}
\label{tab:tap5}
\end{table}

\subsection{Analysis of Losses}
To analyze the impact of the three loss functions  $L_{rank}^n$, $L_{rank}^a$, and ${L_{dil}}$, we perform ablation experiments on the UCF-Crime and XD-Violence datasets. As shown in \cref{tab:tap5}, when all three loss functions are absent, the performance of our method is unsatisfactory. This reveals that the original CLIP suffers from domain bias and is not directly applicable to the VAD domain. When three loss functions are available individually, the performance of our method is clearly improved, where the $L_{rank}^a$ gives the biggest boost to the performance. When all three losses are combined and cooperate with each other, our method achieves the best performance. This demonstrates the effectiveness of the three loss functions we have designed, and they can effectively assist CLIP in domain adaptation for WSVAD.
\section{Conclusions}
In this paper, we propose a novel framework, TPWNG, to perform pseudo-label generation and self-training for WSVAD. TPWNG finetunes CLIP with the designed ranking loss and distributional inconsistency loss to transfer its text-image alignment capability to assist pseudo-label generation with the PLG module. Further, we design a learnable text prompt and normality visual prompt mechanisms to further improve the alignment accuracy of video events description text and video frames. Finally, we introduce a TCSAL module to learn the temporal dependencies of different video events more flexibly and accurately. We perform extensive experiments on the UCF-Crime and XD-Violence datasets, and the superior performance compared to existing methods demonstrates the effectiveness of our method.

\section{Acknowledgments}
This work was supported by the Guangzhou Key Research and Development Program (No. 202206030003),
the Fundamental Research Funds for the Central Universities, the Innovation Fund of Xidian University (No. YJSJ24006), and the Guangdong High-level Innovation Research Institution Project (No. 2021B0909050008).


{
    \small
    \bibliographystyle{ieeenat_fullname}
    \bibliography{main}
}
\clearpage
\setcounter{page}{1}
\maketitlesupplementary











\section{Network Structure Details}
\begin{figure}[t]
  \centering
   \includegraphics[width=0.8\linewidth]{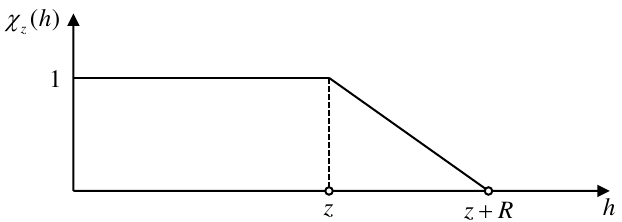}

   \caption{The shape of the soft mask function ${\chi _z}(h)$.}
   \label{fig:hanshu}
\end{figure}

\textbf{TCSAL.} The TCSAL module consists of 4 transformer-encoder layers with 4 attention heads per layer, and each self-attention head of each layer is self-adaptively adjusting its attention span by a soft-masking function ${\chi _z}(h)$. The shape of the soft mask function ${\chi _z}(h)$ is shown in \cref{fig:hanshu}.
\textbf{Classifier.} The classifier adopts a simple structure that consists of a layer normalization layer, a linear layer, and a sigmoid layer.

\section{Finetuning and Prompt Learning}
The finetuning of the CLIP text encoder is performed together with the training of the NVP, PLG, TCSAL, and Classifier modules. During this process, the weights of both the CLIP image and text encoder are frozen, except for the last projection layer of the text encoder which is unfrozen for finetuning. 

To set the optimal fine-tuning configuration, we perform finetuning experiments on the final projection layers of the CLIP image encoder and text encoder as shown in \cref{tab:tap6}. We find that based on prompt learning, there is a large performance improvement after fine-tuning the CLIP text encoder alone, whereas if we finetune only the final projection layer of the CLIP image encoder or both the image encoder and the text encoder at the same time, the performance instead decreases in both cases. Thus our final choice is prompt learning + finetuning (text encoder). We think this is due to the relatively small video anomaly dataset causing overfitting of the CLIP image encoder, which affects the method performance. When finetuning the text encoder alone, the overfitting situation is mitigated because of the prompt learning assistance. Finetuning also facilitates domain adaptation, so this combination of prompt learning + finetuning (text encoder) performs optimally. 

\begin{table}[htbp]
\centering
\scriptsize
\arrayrulecolor{black}
\begin{tabular}{l|c|c} 
\hline
CLIP finetuning and prompt learning configurations                                       & UCF (AUC) & XD (AP)  \\ 
\hline
No finetune
  + prompt learning                                             & 86.45\%   & 81.33\%  \\ 
\arrayrulecolor{black}\hline
Image encoder
  finetuning + prompt learning                                  & 84.23\%   & 81.16\%  \\ 
\hline
Text\&Image encoder finetuning + prompt learning                            & 85.76\%   & 82.12\%  \\ 
\hline
\rowcolor[rgb]{0.753,0.753,0.753} Text encoder
  finetuning + prompt learning & 87.79\%   & 83.68\%  \\
\arrayrulecolor{black}\hline
\end{tabular}
\caption{The AUC and AP change of our method on the UCF-Crime and XD-Violence datasets with different finetuning and prompt learning configurations.}
\label{tab:tap6}
\end{table}

\section{Training and Inference}
Different from existing two-stage methods that separate pseudo-label generation and classifier self-training into two stages, in our approach, we synchronize pseudo-label generation and classifier training until both converge. This ensures that the updated pseudo-labels are used for supervised classifier training in real time, minimizing the interference of noisy labels on classifier training. After training under the supervision of the generated pseudo-labels, only the CLIP image encoder, the TCSAL, and the classifier are involved in the testing phase, where the video frame anomaly scores are predicted directly by the classifier.

\section{Implementation Details.}
Our method is implemented on a single NVIDIA RTX 3090 GPU using the Pytorch framework. We use Adam optimizer with a weight decay of 0.005. The batch size is set to 64, which contains 32 normal videos and 32 abnormal videos randomly sample from the training dataset. For the UCF-Crime dataset, the learning rate and total epoch are set to 0.001 and 50, respectively. For the XD-Violence dataset, the learning rate and the total epoch are set to 0.0001 and 20, respectively. 

\section{Impact of Normality Guidance Weight $\alpha$.}
\begin{figure}[t]
  \centering
   \includegraphics[width=0.75\linewidth]{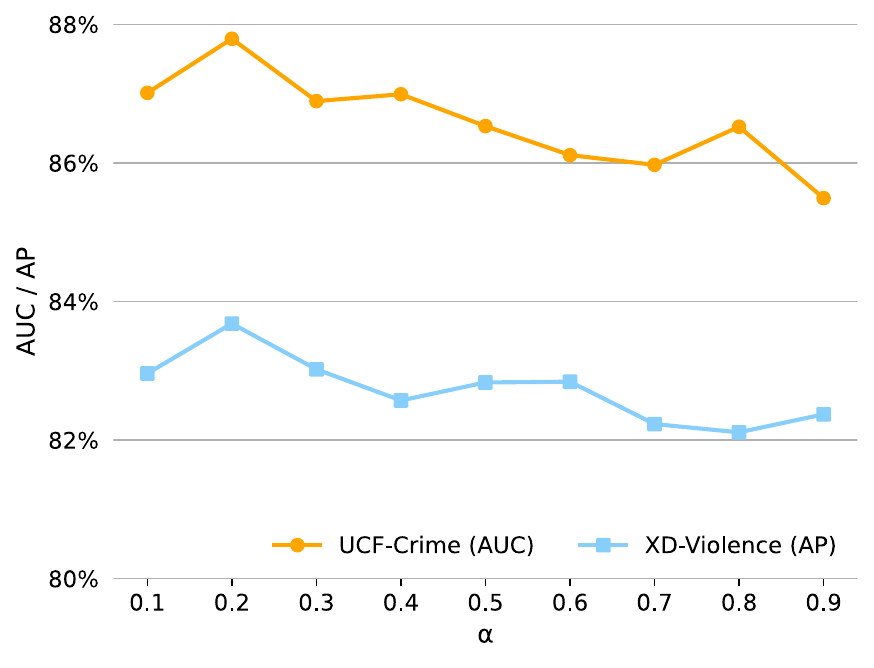}

   \caption{The AUC and AP change of our method on the UCF-Crime and XD-Violence datasets with different normality guidance weight $\alpha$.}
   \label{fig:alpha}
\end{figure}

The normality guidance weight $\alpha$ is used to control the degree of fusion of $\tilde S_{i,\;k}^{an}$ and $\tilde S_{i,\;\tau }^{aa}$ during pseudo-labels generation. In order to analyze the effect of $\alpha$, we set different values of $\alpha$ for comparison experiments. As shown in \cref{fig:alpha}, our method achieves optimal performance on both UCF-Crime and XD-Violence datasets when $\alpha$ is set to 0.2. It can be observed that as $\alpha$ gradually increases, the performance of our method gradually decreases, we consider that it is because too large $\alpha$ instead affects the alignment of the real anomaly event description text and the anomaly frames, and $\alpha=0.2$ is the best trade-off.

\section{Impact of Pseudo-label Generation Threshold $\theta$.}
\begin{figure}[t]
  \centering
   \includegraphics[width=0.75\linewidth]{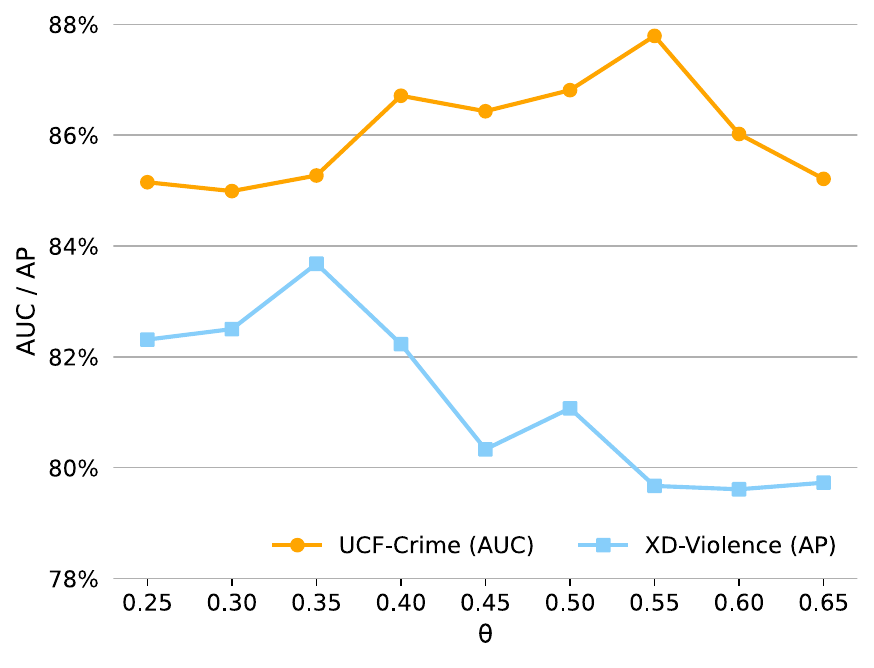}

   \caption{The AUC and AP change of our method on the UCF-Crime and XD-Violence datasets with different pseudo-label generation threshold $\theta$.}
   \label{fig:theta}
\end{figure}

To analyze the impact of different pseudo-label generation thresholds $\theta$ on the performance of our method, we set up a series of different thresholds $\theta$ to perform comparative experiments. As shown in \cref{fig:theta}, the two datasets have different sensitivities to the threshold $\theta$. When $\theta$ is set to 0.55 and 0.35, our method achieves the optimal performance on UCF-Crime and XD-Violence datasets, respectively.

\section{Impact of Context Length $l$ in Learnable Prompt.} 
To investigate the optimal length of learnable textual prompts, we conduct comparative experiments with the context length $l$ being set to 4, 8, 16, and 32, respectively. As shown in \cref{tab:tap7}, both datasets achieve the best performance with the context length $l$ set to 8, and slightly lower performance with a length of 16. However, when the context length $l$ is set to 4 or 32, the performance of our method suffers a large degradation. We conjecture that the reason for this result is that too short a context length leads to textual prompts that do not fully characterize the video frame events, leading to model underfitting. Conversely, too long context length may lead to model overfitting.

\begin{table}
\centering
\arrayrulecolor{black}
\begin{tabular}{c!{\color{black}\vrule}c!{\color{black}\vrule}c} 
\hline
$l$ & UCF-Crime (AUC) & XD-Violence (AP)  \\ 
\cline{1-2}\arrayrulecolor{black}\cline{3-3}
4              & 82.26\%           & 77.45\%             \\ 
\arrayrulecolor{black}\cline{1-2}\arrayrulecolor{black}\cline{3-3}
8              & \textbf{87.79\%}  & \textbf{83.68\%}    \\ 
\arrayrulecolor{black}\cline{1-2}\arrayrulecolor{black}\cline{3-3}
16             & 87.24\%           & 82.99\%             \\ 
\arrayrulecolor{black}\cline{1-2}\arrayrulecolor{black}\cline{3-3}
32             & 85.23\%           & 81.78\%             \\
\arrayrulecolor{black}\cline{1-2}\arrayrulecolor{black}\cline{3-3}
\end{tabular}
\arrayrulecolor{black}
\caption{The AUC and AP of our method on the UCF-Crime and XD-Violence datasets with different context lengths $l$.}
\label{tab:tap7}
\end{table}

\begin{figure}[t]
  \centering
   \includegraphics[width=0.85\linewidth]{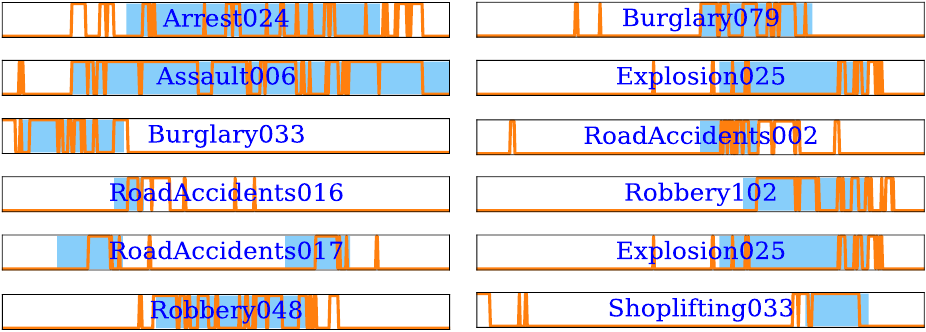}
    \caption{Visualization of pseudo-labels of some video clips on the UCF-Crime dataset.}
     \label{fig:pl}
\end{figure}

\section{Visualization of Pseudo-labels.}
We visualize part of the pseudo-labels (UCF-Crime) in \cref{fig:pl}. The generated pseudo-labels (orange solid line) approximate the ground-truth (shades of blue) well in most cases, which indicates the effectiveness of the generated pseudo-labels.

\section{Visualization of Match Similarities.}
\begin{figure*}[t]
  \centering
   \includegraphics[width=1.0\linewidth]{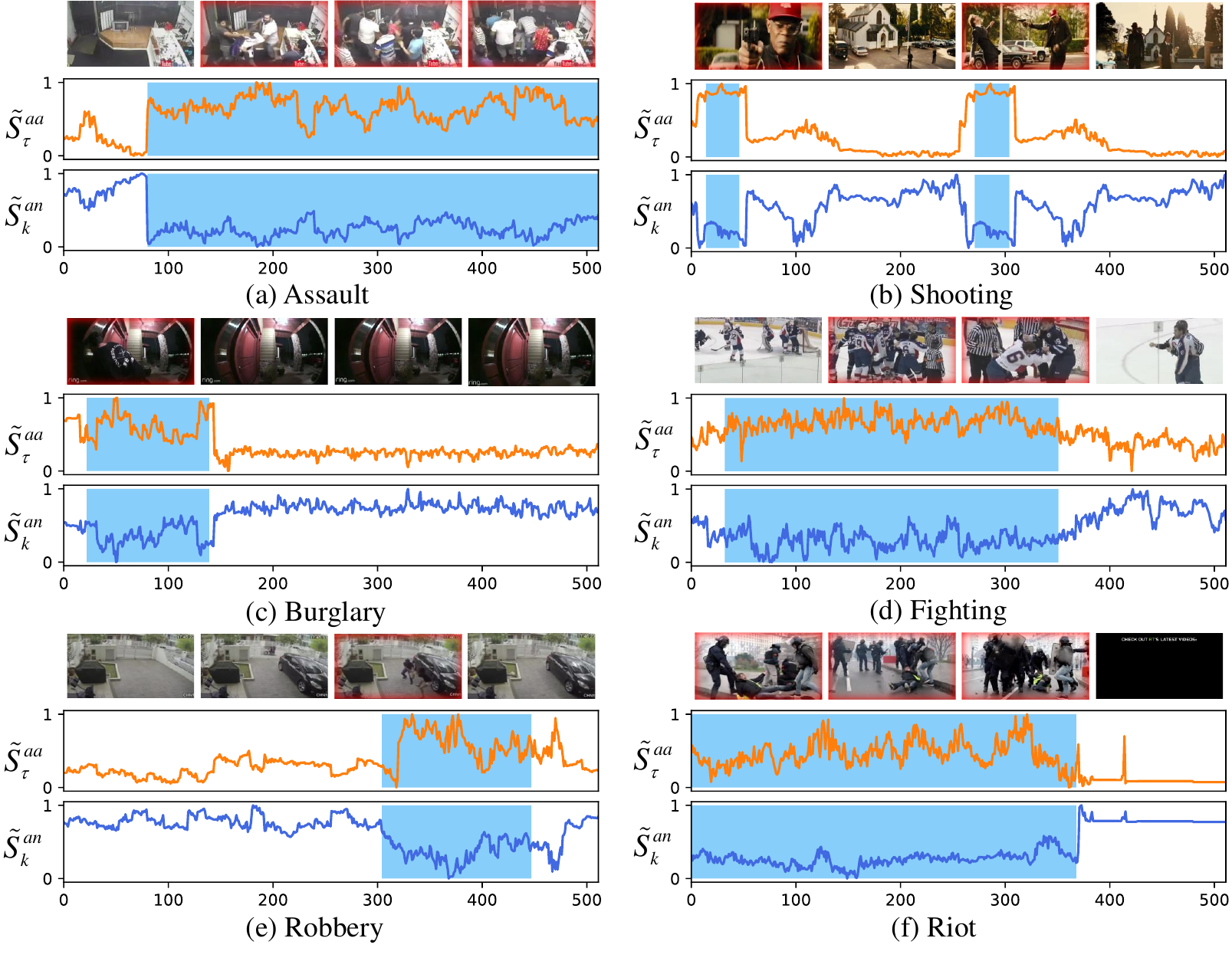}

   \caption{Visualization of match similarities between video event description text and video frames for several anomaly samples on the UCF-Crime and XD-Violence test datasets. The light blue range represents abnormal ground truth.}
   \label{fig:smi}
\end{figure*}

To more intuitively show that our constructed framework can facilitate the CLIP model to perform domain adaptation for matching video event text descriptions and corresponding video frames, we visualize the $\tilde S_{\tau}^{aa}$ and $\tilde S_{k}^{an}$, i.e., the match similarities of real abnormal event description text and normal event description text with corresponding abnormal videos, on the UCF-Crime and XD-Violence datasets, respectively. We can observe from \cref{fig:smi} that the distributions of $\tilde S_{\tau}^{aa}$ and $\tilde S_{k}^{an}$ are contradictory which can align anomalous video frames and normal video frames, respectively. This shows the effectiveness of our designed distributional inconsistency loss $L_{dil}$. In addition, we can notice from \cref{fig:smi} (a) and (f) that there are fluctuations in the alignment of the real abnormal event description text and the corresponding abnormal video frames in these two samples, while the normal event description text has a more accurate alignment, in which case our proposed normal guidance mechanism can assist $\tilde S_{\tau}^{aa}$ to better align the abnormal video frames.


\end{document}